\ifdefined\XeTeXversion\else
  \ifdefined\pdfoutput\pdfoutput=1\fi
\fi
\documentclass[runningheads]{llncs}

\usepackage{eccv}

\usepackage[width=122mm,left=12mm,paperwidth=146mm,height=193mm,top=12mm,paperheight=217mm]{geometry}

\usepackage{eccvabbrv}

\usepackage{graphicx}
\usepackage{booktabs}
\usepackage{multirow}

\usepackage{hyperref}

\usepackage{orcidlink}

\begin{document}

\title{Open-Weather Robust 3D Detection via Dual-Critic Diffusion Alignment
}

\titlerunning{Open-Weather Robust 3D Detection}

\author{
Shuyao Li\inst{1}\orcidlink{0009-0007-0456-1007} \and
Chuanxing Geng\inst{1,2}\orcidlink{0000-0001-6345-5385}\thanks{Corresponding author.} \and
Heyang Sun\inst{1} \and
Qiang Zhou\inst{1}\orcidlink{0000-0002-4577-0581} \and
Jingjing Gu\inst{1}\orcidlink{0000-0002-3989-1520}
}

\authorrunning{S.~Li et al.}

\institute{
Nanjing University of Aeronautics and Astronautics, Nanjing, China\\
\email{\{lsyaoo, gengchuanxing, zhouqnuaacs, gujingjing\}@nuaa.edu.cn}
\and
MIIT Key Laboratory of Pattern Analysis and Machine Intelligence, Nanjing, China
}
\maketitle

\begin{abstract}
Robust 3D object detection under adverse weather remains a critical hurdle for autonomous driving. Despite progress with LiDAR--4D radar fusion, most methods are constrained by a closed-world assumption, implicitly requiring training and test weather to align in both type and severity. This premise fails in practice: the open-ended nature of weather, and even variations within a single type like rain, cause dramatically different LiDAR degradation patterns, leading to significant performance drops in unseen conditions. To address this, we present Dual-Critic Guided Diffusion Alignment (DCDA), a \textit{weather-agnostic} framework that learns to recover degraded LiDAR features toward a clean manifold. Rather than modeling specific weather types, DCDA employs a 4D radar-conditioned diffusion process to progressively refine features, guided by two complementary critics. (i) A detection-guided critic, anchored by a pre-trained clean-weather model, ensures that the refined features retain object-level discriminability and localization accuracy. (ii) A weather adversarial critic enforces holistic distributional consistency with clean-weather representations. By aligning features through semantic and distributional constraints rather than explicit weather modeling, DCDA generalizes effectively to unseen weather types and severities without requiring paired data or weather labels. We further introduce a structured open-weather benchmark with held-out type--severity combinations and extensive experiments verify DCDA's advantages. Code is available at \url{https://github.com/Mangonn/DCDA}.

\end{abstract}

\section{Introduction}
\label{sec:introduction}

\begin{figure}[t]
    \centering
    \includegraphics[width=\linewidth]{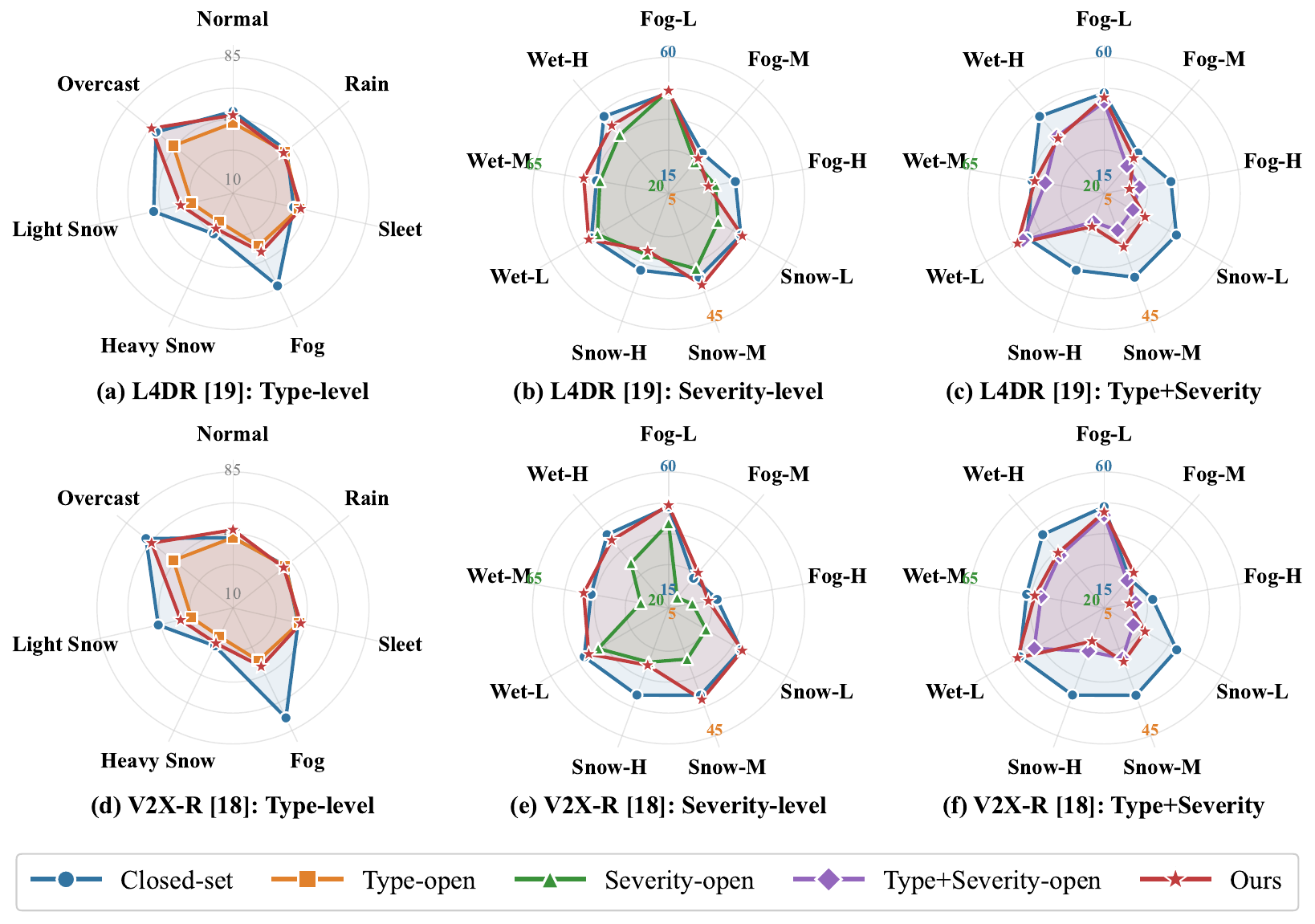}
    \caption{
Open-weather generalization gap on K-Radar.
The top and bottom rows show BEV AP for L4DR and V2X-R, respectively, under the three protocols shown from left to right: \emph{type-open}, \emph{severity-open}, and \emph{type+severity-open}.
\emph{Closed-set} denotes each detector's conventional closed-set baseline, where the test weather conditions are seen during training.
\textbf{Ours} denotes the proposed DCDA model under the corresponding open-weather protocol.
Type-open holds out real weather types, severity-open holds out synthetic severity levels, and type+severity-open holds out both synthetic weather type and severity.
The drop from \emph{Closed-set} to the open-weather curves reveals a clear generalization gap for the two representative baselines.
}

    \label{fig:motivation}
\end{figure}

Robust 3D object detection is fundamental to autonomous driving, where perception errors directly impact planning and safety-critical decisions \cite{MultiModalSurvey2022}.
While modern LiDAR-based detectors achieve strong performance in clear conditions, their reliability degrades sharply under adverse weather such as fog, rain, and snow \cite{Bijelic2020SeeingThroughFog,Hahner2021FogSim,Hahner2022SnowSim}. Weather-induced degradation is inherently asymmetric across sensing modalities. Fog and snow introduce severe backscatter and range attenuation in LiDAR, corrupting geometric structure \cite{Bijelic2020SeeingThroughFog}. In contrast, millimeter-wave radar remains comparatively robust and preserves long-range and Doppler cues \cite{Chae2024_3DLRF,l4dr,ding2024radarocc}. Motivated by this complementarity, recent work increasingly adopts LiDAR--4D radar fusion \cite{Paek2022KRadar,l4dr}.

However, despite their improved robustness, most existing fusion methods \cite{Chae2024_3DLRF,l4dr,v2xr,qi2026fusionbev,wang2022interfusion} implicitly rely on a closed-world assumption: the weather conditions encountered at test time are assumed to be represented during training or restricted to predefined corruption settings \cite{robo3d,dong2023benchmarking}. This assumption rarely holds in real deployment. Weather is open-ended and continuously varying in both type and severity. Even within a single category such as snow or fog, intensity variations can induce substantially different LiDAR degradation patterns \cite{Hahner2021FogSim,Hahner2022SnowSim,wu2025weathergen}. As a result, models trained under limited weather configurations may fail when exposed to unseen type--severity combinations.

To examine how current state-of-the-art fusion detectors behave under such open-weather conditions, we construct a structured benchmark that jointly considers variations in weather type and severity. Based on K-Radar \cite{Paek2022KRadar}, we design systematic holdout protocols over (i) weather types, (ii) severity levels, and (iii) their combinations, simulating unseen environmental configurations at test time. Since K-Radar provides limited severity annotations, we further construct controlled intensity levels (Light/Medium/Heavy) by progressively synthesizing weather effects on \textit{Normal} samples~\cite{robo3d,wu2025weathergen}, enabling systematic robustness evaluation across severity variations.

As shown in Fig.~\ref{fig:motivation}, our findings are striking: Both representative state-of-the-art baselines, L4DR \cite{l4dr} and V2X-R \cite{v2xr} exhibit consistent and often substantial performance degradation under these open settings compared to conventional closed-world evaluations. This reveals a clear open-weather generalization gap that has not been adequately addressed. These observations suggest that robustness cannot rely solely on enumerating weather types or simulating predefined degradations. Instead, models must learn weather-agnostic mechanisms capable of recovering stable geometric representations beyond seen conditions. 

To this end, we propose Dual-Critic Guided Diffusion Alignment (DCDA), a weather-agnostic framework that restores degraded LiDAR features by aligning them to a clean (weather) manifold. Rather than explicitly modeling specific weather patterns, DCDA formulates feature recovery as a 4D radar-conditioned diffusion process, progressively refining corrupted LiDAR features \cite{zou2024diffbev,ye2025bevdiffuser,chen2024diffubox,v2xr}. A central challenge is to enhance degraded features while mitigating negative transfer from imperfect radar cues and preserving detection semantics. We address this using two complementary critics that regulate the diffusion trajectory:
\begin{itemize}
    \item[$\checkmark$] \textbf{A detection-guided critic}, instantiated by a frozen \textit{Normal} detector to enforce task-level semantic consistency and localization fidelity;
    \item[$\checkmark$] \textbf{A weather adversarial critic}, which enforces global distributional alignment with \textit{Normal} representations.
\end{itemize}
By jointly constraining task-space and representation-space alignment---without explicit weather modeling, paired clean--degraded supervision, or conditioning on specific weather type/severity labels---DCDA generalizes effectively to unseen weather types and severity levels. As shown in Fig.~\ref{fig:motivation}, rather than over-optimizing for specific conditions, our DCDA achieves stronger overall open-weather generalization than both representative baselines on K-Radar.
Averaged over the plotted weather, DCDA improves mean 3D AP over the better baseline by 5.2 percentage points under held-out weather types (type-open),
3.0 percentage points under held-out severity levels (severity-open),
and 1.8 percentage points under their combinations (type+severity-open).

Our contributions are threefold:
\begin{itemize}
    \item We identify and formalize the open-weather generalization problem in LiDAR--4D radar fusion, and construct a structured benchmark that jointly evaluates weather type and severity variations, revealing a substantial robustness gap in existing state-of-the-art detectors.
    \item We propose a weather-agnostic diffusion alignment paradigm (DCDA) that formulates degraded LiDAR feature recovery as a radar-conditioned manifold alignment process, avoiding explicit weather modeling, paired clean--degraded supervision, and weather type/severity labels as conditional inputs.
    \item We introduce a dual-critic regularization mechanism that stabilizes diffusion-based alignment by jointly enforcing task-space semantic consistency and representation-space distribution matching, enabling robust generalization to unseen weather types and severity levels.
\end{itemize}

\section{Related Work}
\label{sec:related_work}
\subsection{LiDAR-Based Robust 3D Object Detection in Adverse Weather}

Adverse weather (e.g., fog, rain, and snow) introduces modality-specific degradations (e.g., attenuation, backscatter clutter, and point sparsification) that corrupt geometric evidence and substantially degrade 3D detection performance \cite{Bijelic2020SeeingThroughFog,Hahner2021FogSim,Hahner2022SnowSim,dong2023benchmarking}.
Prior efforts improve robustness through three directions: 
(i) LiDAR restoration and filtering, including deweathering/denoising and clutter suppression (e.g., snow/rain outlier filters and adverse-effect detection) \cite{huang2023lidsor,piroli2023energy}; 
(ii) robust learning, which enhances adverse-weather generalization via robust objectives (e.g., cross-weather distillation or geometry/structure constraints) \cite{qi2024geometric,wu2026pasenet}; 
and (iii) adaptation and post-processing, which mitigates test-time shifts via deployment-time adaptation or calibrated post-processing \cite{yuan2024reg,chen2024mos,lee2024toward}.
When complementary camera cues are available, degradation-aware camera-LiDAR fusion exploits asymmetric corruption via uncertainty-guided feature exchange and fog-aware depth completion \cite{Bijelic2020SeeingThroughFog,zhang2025fogfusion,zhang2024enhancing}. Context-gated fusion further improves robustness across day/night and clear/rainy conditions \cite{sural2024contextualfusion}.
However, robustness often remains limited once LiDAR geometry is heavily corrupted, motivating the use of weather-tolerant millimeter-wave radar signals.

\subsection{LiDAR--4D Radar Fusion for Robust 3D Detection in Adverse Weather}
Millimeter-wave radar is generally less affected by scattering media than LiDAR~\cite{ding2024radarocc,Peng20254RadarSurvey,Han2023Survey4DRadar}, motivating LiDAR--4D radar fusion for adverse-weather 3D detection~\cite{wang2023bi,Chae2024_3DLRF,l4dr,v2xr}.
K-Radar provides a large-scale benchmark with synchronized LiDAR and 4D radar across diverse real-world weather, enabling systematic study of such fusion~\cite{Paek2022KRadar}.
Recent methods improve robustness via (i) 3D-centric fusion with weather-aware modulation (e.g., 3D-LRF with weather-conditional gating)~\cite{Chae2024_3DLRF}, and (ii) denoising and gated fusion to handle sensor gaps and varying degradation (L4DR)~\cite{l4dr}.
Beyond fusion operator design, radar has also been used as auxiliary supervision to improve LiDAR-only robustness~\cite{chae2024lidar}.
Other LiDAR--4D radar fusion lines explore sensor-adaptive multi-modal fusion~\cite{palladin2024samfusion} and interaction-based attention with adaptive weighting, leveraging radar-specific cues (e.g., velocity/RCS), robustness to sensor failures, and multi-frame radar accumulation with motion-aware compensation~\cite{wang2022interfusion,rlnet,mutualforce,moral,chae2025doppler}.
Diffusion-based denoising further leverages radar as a condition to suppress weather noise in LiDAR features~\cite{v2xr}.
Despite these advances, many existing evaluations follow standard splits where weather categories overlap between training and test, leaving explicit generalization to held-out weather types/severity less explored.

\subsection{Out-of-Distribution (OOD) Weather Generalization}
A limited but growing body of work explicitly targets open-weather generalization and evaluates robustness under controlled OOD corruptions and severity levels~\cite{dong2023benchmarking,robo3d}.
Representative approaches leverage uncertainty-aware mechanisms to handle asymmetric degradations (e.g., SeeingThroughFog)~\cite{Bijelic2020SeeingThroughFog}.
Diffusion-based weather adaptation has also been explored for robust 3D detection; for example, MonoWAD enhances features with a weather-adaptive diffusion model guided by weather-reference representations learned from clear/fog counterparts~\cite{oh2024monowad}.
Recent studies further investigate diffusion for adverse-weather restoration and augmentation beyond monocular settings~\cite{he2025diffusion,matteazzi2025augmentation}.
In parallel, robustness under domain shift has been advanced via unsupervised domain adaptation~\cite{peng2023cl3d,chang2024cmda} and test-time adaptation~\cite{dpo,yuan2024reg,chen2024mos}.
Simulation-free robustness training has also been explored to improve weather-based OOD generalization without relying on costly simulators~\cite{batten2025improving}.

Despite emerging OOD evaluations and adaptation/restoration techniques, two limitations remain for practical LiDAR--4D radar fusion under open-weather deployment.
First, several robustness pipelines still rely on explicit weather priors (e.g., domain identifiers or predefined corruption families) and/or additional signals such as weather annotations or paired clear-adverse references, which are costly to curate and not consistently available across diverse operating conditions.
Second, robustness is commonly characterized under either type shift or severity variation in isolation, while generalization under their joint type--severity shifts is less systematically examined in LiDAR--4D radar fusion settings.
These limitations motivate robustness mechanisms and evaluation protocols that operate under minimal supervision and explicitly account for joint shifts.

\begin{figure}[t]
    \centering
    \includegraphics[width=\linewidth]{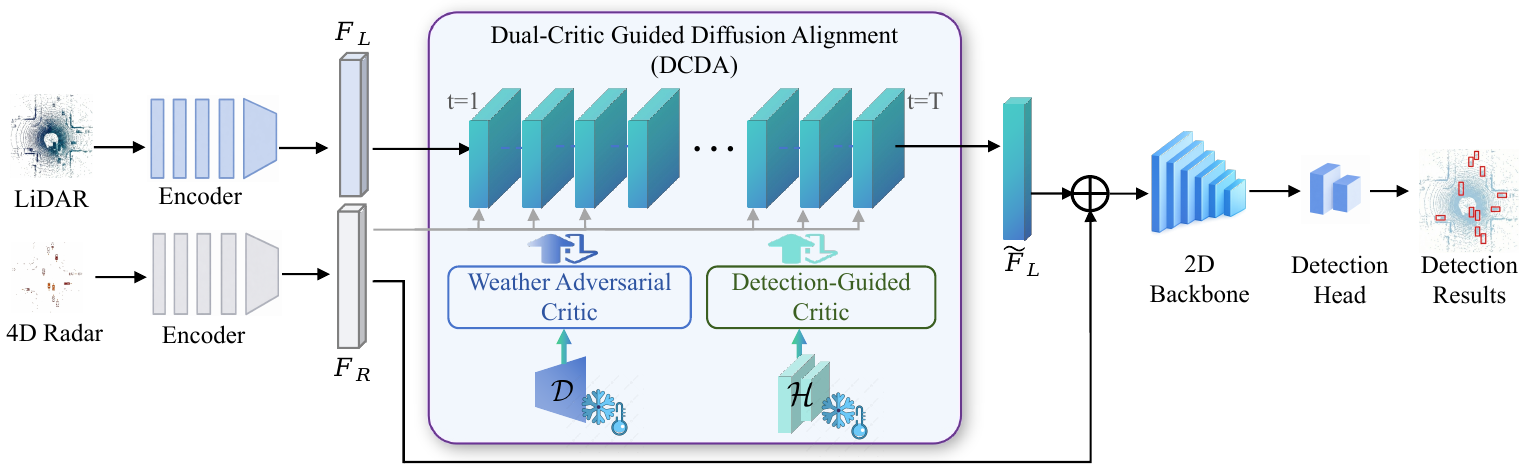}
    \caption{Overview of DCDA. The radar-conditioned diffusion alignment refines LiDAR BEV features, guided by a frozen detection critic and a frozen normal-weather discriminator.}
    \label{fig:train}
\end{figure}

\section{Method}
\label{sec:method}
We study open-weather generalization for LiDAR--4D radar 3D object detection, where test-time conditions may deviate from training in both weather \emph{type} and \emph{severity}.
To sidestep explicit weather modeling and reliance on paired clean-degraded supervision, we pursue weather-agnostic feature recovery, restoring stable LiDAR geometric representations beyond seen conditions.
Specifically, we propose \textbf{Dual-Critic Guided Diffusion Alignment (DCDA)}, which refines the LiDAR BEV feature $\mathbf{F}_L$ into $\tilde{\mathbf{F}}_L$ conditioned on the 4D radar feature $\mathbf{F}_R$ as a comparatively stable signal.
DCDA consists of (i) a \textbf{radar-conditioned diffusion alignment} $\mathcal{A}_\theta$ for iterative refinement, and (ii) a \textbf{dual-critic guidance} toward the clean manifold during training: a detection-guided critic $\mathcal{H}$ that enforces semantic and localization fidelity, and a weather adversarial critic $\mathcal{D}$ that promotes alignment to \textit{Normal}-weather representations.
At inference, $\mathcal{D}$ can be optionally reused as a lightweight router to bypass refinement for confidently \textit{Normal} inputs.
An overview is shown in Fig.~\ref{fig:train}.

\subsection{Radar-Conditioned Diffusion Alignment}
\label{sec:diffusion}

We model feature refinement as a radar-conditioned diffusion process operating on LiDAR BEV features.
Given the input LiDAR feature $\mathbf{F}_0 \equiv \mathbf{F}_L$ (possibly degraded),
the forward diffusion process progressively corrupts it with Gaussian noise over $T$ steps:
\begin{equation}
\mathbf{F}_t
=
\sqrt{\bar{\alpha}_t}\,\mathbf{F}_0
+
\sqrt{1-\bar{\alpha}_t}\,\boldsymbol{\epsilon},
\quad
\boldsymbol{\epsilon}\sim\mathcal{N}(\mathbf{0},\mathbf{I}),
\end{equation}
where $\bar{\alpha}_t=\prod_{i=1}^{t}\alpha_i$ follows a predefined variance schedule.

Unlike generative diffusion models that start from pure Gaussian noise,
we initialize the reverse chain from a noised observation
$\mathbf{F}_T\sim q(\mathbf{F}_T\mid \mathbf{F}_0)$.
This retains the coarse structure of the input and confines denoising to a local neighborhood around $\mathbf{F}_0$,
serving as a conditional refiner.

At each reverse step, a radar-conditioned U-Net $\mathcal{A}_\theta$ directly predicts a \textit{Normal}-weather feature estimate:
\begin{equation}
\hat{\mathbf{F}}_0^{(t)} = \mathcal{A}_\theta([\mathbf{F}_t;\,\mathbf{F}_R],\,t).
\end{equation}
This prediction is used to compute the reverse transition via the standard DDPM posterior:
\begin{equation}
\mu_\theta
=
\frac{\sqrt{\bar{\alpha}_{t-1}}\,\beta_t}{1-\bar{\alpha}_t}\,\hat{\mathbf{F}}_0^{(t)}
+
\frac{\sqrt{\alpha_t}\,(1-\bar{\alpha}_{t-1})}{1-\bar{\alpha}_t}\,\mathbf{F}_t,
\end{equation}
where $\beta_t=1-\alpha_t$ and the posterior variance is
$\tilde{\beta}_t=\frac{(1-\bar{\alpha}_{t-1})}{1-\bar{\alpha}_t}\beta_t$.
After $T$ reverse steps, we obtain the refined feature $\tilde{\mathbf{F}}_L=\hat{\mathbf{F}}_0^{(0)}$,
which replaces the original $\mathbf{F}_L$ for downstream fusion.

We apply a self-reconstruction loss to maintain the geometric structure of the input features:
\begin{equation}
\mathcal{L}_{\text{diff}}
=
\,\left\|
\tilde{\mathbf{F}}_L-\mathbf{F}_L
\right\|_2^2.
\end{equation}
It serves as a content-preserving regularizer to stabilize optimization and prevent over-correction, while the alignment toward the clean manifold is driven by the dual critics, as described in Sec.~\ref{sec:critics}.

\subsection{Dual-Critic Guidance Toward the Clean Manifold}
\label{sec:critics}
We construct two \emph{frozen} critics to steer diffusion alignment toward the clean manifold:
(1) a \textbf{detection-guided critic} $\mathcal{H}$, trained on \textit{Normal}-only data and then fixed to provide \textit{Normal}-domain semantic guidance;
and (2) a \textbf{weather adversarial critic} $\mathcal{D}$, trained to distinguish \textit{Normal} vs.\ non-\textit{Normal} LiDAR BEV features and then fixed to enforce distributional alignment to the \textit{Normal}-domain.

\par\medskip\noindent\textbf{Detection-guided critic (semantic constraint).}
To preserve object-level semantics, we reuse the standard detection loss computed by the frozen head $\mathcal{H}$:
\begin{equation}
\mathcal{L}_{\text{det}}
=
\mathcal{L}_{\text{cls}}
+
\mathcal{L}_{\text{loc}}.
\end{equation}
Since $\mathcal{H}$ is pretrained on only \textit{Normal} data, minimizing $\mathcal{L}_{\text{det}}$ requires $\tilde{\mathbf{F}}_L$ to remain in regions of the feature space that the \textit{Normal} detector recognizes as object-discriminative,
providing a task-driven \emph{semantic pull} toward the clean manifold.

\par\medskip\noindent\textbf{Weather adversarial critic (distributional constraint).}
We enforce distributional consistency with clean features via $\mathcal{D}(\cdot)$ on LiDAR BEV features.
Let $\mathbf{F}_L^{\text{N}}$ and $\mathbf{F}_L^{\text{non-N}}$ denote LiDAR BEV features from the \textit{Normal} and non-\textit{Normal} training samples, respectively.
We train the discriminator $\mathcal{D}$ to predict whether a feature comes from the \textit{Normal} subset using the binary cross-entropy loss:
\begin{equation}
\mathcal{L}_\mathcal{D}
=
-\mathbb{E}_{\mathbf{F}_L^{\text{N}}}
\big[\log \mathcal{D}(\mathbf{F}_L^{\text{N}})\big]
-
\mathbb{E}_{\mathbf{F}_L^{\text{non-N}}}
\big[\log \big(1 - \mathcal{D}(\mathbf{F}_L^{\text{non-N}})\big)\big].
\end{equation}
After pretraining, $\mathcal{D}$ is kept frozen and used to define an adversarial guidance term for refined features:
\begin{equation}
\mathcal{L}_{\text{adv}}
=
-\mathbb{E}_{\tilde{\mathbf{F}}_L}
\big[\log \mathcal{D}(\tilde{\mathbf{F}}_L)\big].
\end{equation}

\par\medskip\noindent\textbf{Critic-guided refinement.}
We apply both critics on the refined LiDAR feature $\tilde{\mathbf{F}}_L$ produced by the diffusion aligner $\mathcal{A}_\theta$ and form
\begin{equation}
\mathcal{L}_{\text{crit}}
=
\lambda_1 \mathcal{L}_{\text{det}}
+
\lambda_2 \mathcal{L}_{\text{adv}},
\label{eq:lcrit}
\end{equation}
whose gradients are backpropagated only to $\mathcal{A}_\theta$ while keeping $\mathcal{H}$ and $\mathcal{D}$ frozen.
$\mathcal{L}_{\text{crit}}$ is combined with the diffusion objective to form the overall training loss, as described in Sec.~\ref{sec:training_inference} and formalized in Eq.~\eqref{eq:totalloss}.
These \textit{Normal}-domain critics provide complementary semantic and distributional guidance, regularizing the refinement and mitigating potential drift.

\subsection{Training and Inference}
\label{sec:training_inference}
\par\medskip\noindent\textbf{Training.}
Optimizing the aligner $\mathcal{A}_\theta$ proceeds in two stages, while keeping both critics $\mathcal{H}$ and $\mathcal{D}$ frozen.

\noindent\textit{Stage I: normal weather prior warm-up.}
We train $\mathcal{A}_\theta$ on \textit{Normal} samples to learn an identity-preserving refinement under radar conditioning and to stabilize optimization:
\begin{equation}
\mathcal{L}_{\text{I}}
=
\mathcal{L}_{\text{diff}}
+
\lambda_1 \mathcal{L}_{\text{det}}.
\end{equation}

\noindent\textit{Stage II: adversarial alignment.}
We fine-tune $\mathcal{A}_\theta$ on the full training set $\mathcal{W}_{\text{train}}$.
The discriminator term encourages refined features to match the \textit{clean}-domain distribution, while the detection critic preserves task semantics:
\begin{equation}
\mathcal{L}_{\text{total}}
=
\mathcal{L}_{\text{diff}}
+
\lambda_1 \mathcal{L}_{\text{det}}
+
\lambda_2 \mathcal{L}_{\text{adv}}.
\label{eq:totalloss}
\end{equation}
Over Stage~II, the weight on $\mathcal{L}_{\text{diff}}$ is annealed toward $0$, so the frozen critics increasingly drive the alignment.

\par
\begin{figure}[t]
    \centering
    \includegraphics[width=\linewidth]{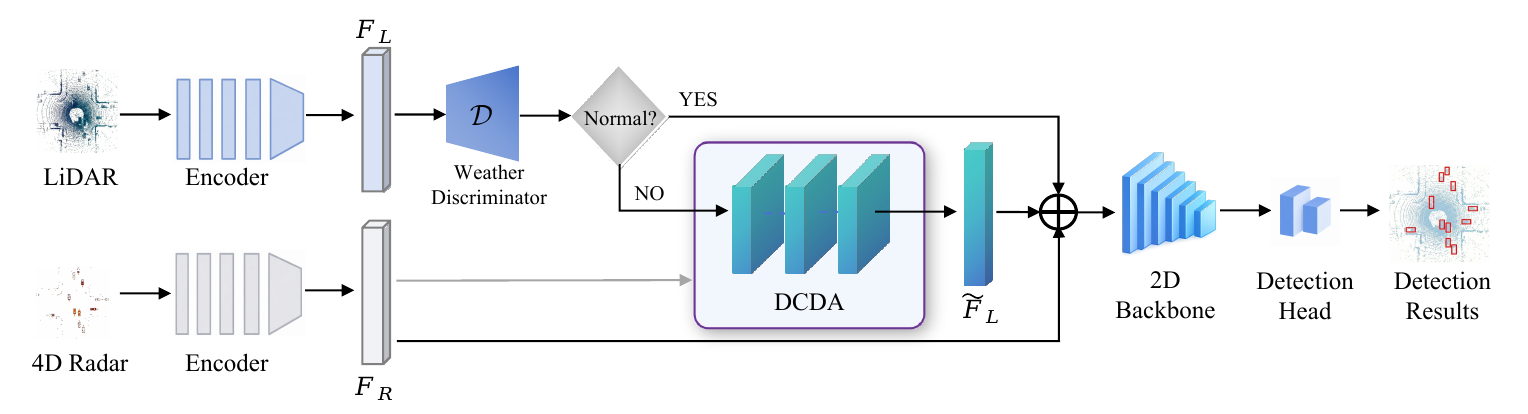}
    \caption{Inference of DCDA with optional routing. When the input is confidently \textit{Normal}, DCDA can be bypassed to reduce unnecessary refinement.}
    \label{fig:inference}
\end{figure}

\medskip\noindent\textbf{Inference.}
As shown in Fig.~\ref{fig:inference}, we reuse the discriminator $\mathcal{D}$ as a lightweight router to decide whether to activate DCDA.
Let $s=\mathcal{D}(\mathbf{F}_L)\in[0,1]$ denote the confidence of \textit{Normal}.
We activate DCDA only when $s<\tau$ and otherwise bypass it:
\begin{equation}
\mathbf{F}_L^{\text{route}}=
\begin{cases}
\mathbf{F}_L, & s\ge\tau \quad (\text{bypass DCDA}),\\
\tilde{\mathbf{F}}_L, & s<\tau \quad (\text{activate DCDA}),
\end{cases}
\end{equation}
where $\tilde{\mathbf{F}}_L$ is the aligned LiDAR feature produced by DCDA when activated.
The routed LiDAR features $\mathbf{F}_L^{\text{route}}$ are then fused with radar features and fed into the downstream detection network.
We set $\tau$ on a held-out split from the training-weather domain without using any test data. Pseudo-code is provided in Sec.~B of the supplementary material.

\section{Experiments}
\label{sec:experiments}

\begin{table}[t]
\centering
\caption{Results on K-Radar \textbf{type-open}: per-weather BEV and 3D AP (\%). Best and second-best are \textbf{bold} and \underline{underlined}.}
\label{tab:kradar_type_open}
\setlength{\tabcolsep}{3.2pt}
\scriptsize
\begin{tabular}{l ccc cccc}
\toprule
\multirow{2}{*}{Method}
& \multicolumn{3}{c}{Seen}
& \multicolumn{4}{c}{Unseen} \\
\cmidrule(lr){2-4}\cmidrule(lr){5-8}
& Normal & Rain & Sleet
& Fog & HeavySnow & LightSnow & Overcast \\
\midrule

\multicolumn{8}{c}{\textbf{BEV AP (\%)}} \\
\midrule

RTNH~\cite{Paek2022KRadar}
& 35.64 & 38.50 & 30.41 & 72.43 & 31.63 & 61.93 & 64.96 \\

InterFusion~\cite{wang2022interfusion}
& 67.65 & 66.71 & 56.93 & 62.05 & 24.36 & 56.40 & 71.22 \\

SpikingRTNH~\cite{spikingrtnh}
& 30.48 & 32.75 & 29.75 & 66.72 & 31.27 & 47.66 & 56.25 \\

V2X-R~\cite{v2xr}
& 65.96 & \underline{68.00} & \underline{57.46} & 69.43 & 30.37 & 66.72 & 78.99 \\

L4DR~\cite{l4dr}
& \underline{68.33} & \textbf{68.16} & 56.02 & \underline{81.32} & \underline{31.74} & \underline{74.81} & \underline{80.42} \\

DCDA (Ours)
& \textbf{68.39} & 67.15 & \textbf{64.76} & \textbf{88.81} & \textbf{35.94} & \textbf{77.07} & \textbf{81.76} \\

\midrule
\multicolumn{8}{c}{\textbf{3D AP (\%)}} \\
\midrule

RTNH~\cite{Paek2022KRadar}
& 21.69 & 20.98 & 14.78 & 10.36 & 11.12 & 9.79 & 28.57 \\

InterFusion~\cite{wang2022interfusion}
& 42.88 & 41.04 & 39.51 & 34.26 & 11.52 & 28.96 & 40.51 \\

SpikingRTNH~\cite{spikingrtnh}
& 11.50 & 19.03 & 10.46 & 23.40 & 19.23 & 8.07 & 29.96 \\

V2X-R~\cite{v2xr}
& 41.61 & 40.70 & \textbf{47.09} & 28.65 & 20.70 & 21.87 & 36.44 \\

L4DR~\cite{l4dr}
& \underline{44.71} & \textbf{43.60} & 36.46 & \underline{36.73} & \underline{21.88} & \underline{29.60} & \underline{49.85} \\

DCDA (Ours)
& \textbf{49.92} & \underline{41.81} & \underline{44.69} & \textbf{41.91} & \textbf{26.29} & \textbf{35.09} & \textbf{65.70} \\

\bottomrule
\end{tabular}
\end{table}

\begin{table}[t]
\centering
\caption{Results on synthetic \textbf{severity-open}: BEV and 3D AP (\%) on Fog/Snow/Wet Ground at Light/Medium/Heavy levels. Training sees \emph{Normal} and \emph{Light} weather only.}
\label{tab:synthetic_severity_open}
\setlength{\tabcolsep}{3.6pt}
\scriptsize

\begin{tabular}{l ccc ccc ccc}
\toprule
\multirow{2}{*}{Method}
& \multicolumn{3}{c}{Fog}
& \multicolumn{3}{c}{Snow}
& \multicolumn{3}{c}{Wet Ground} \\
\cmidrule(lr){2-4}
\cmidrule(lr){5-7}
\cmidrule(lr){8-10}

& L & M & H
& L & M & H
& L & M & H \\
\midrule

\multicolumn{10}{c}{\textbf{BEV AP (\%)}} \\
\midrule

RTNH~\cite{Paek2022KRadar}
& 39.39 & 30.32 & 34.60
& 33.55 & 26.20 & \underline{33.19}
& 40.88 & 27.25 & 32.55 \\

InterFusion~\cite{wang2022interfusion}
& 60.92 & 39.78 & 34.22
& 28.74 & 32.40 & 24.92
& 74.12 & 65.72 & 66.47 \\

L4DR~\cite{l4dr}
& \underline{65.55} & \underline{44.43} & \underline{35.13}
& \underline{34.16} & \underline{35.98} & \textbf{34.19}
& 73.54 & \underline{67.19} & \underline{67.08} \\

V2X-R~\cite{v2xr}
& 65.45 & 39.53 & 32.20
& 28.16 & 34.78 & 30.96
& \textbf{79.50} & 62.44 & 64.43 \\

DCDA (Ours)
& \textbf{68.81} & \textbf{44.70} & \textbf{35.16}
& \textbf{42.35} & \textbf{44.27} & 32.77
& \underline{75.13} & \textbf{67.40} & \textbf{67.37} \\

\midrule
\multicolumn{10}{c}{\textbf{3D AP (\%)}} \\
\midrule

RTNH~\cite{Paek2022KRadar}
& 21.66 & 11.97 & 17.13
& 19.50 & 17.42 & 15.49
& 15.09 & 17.10 & 16.91 \\

InterFusion~\cite{wang2022interfusion}
& \underline{45.94} & 19.04 & 22.86
& 19.01 & 21.61 & 15.21
& 41.89 & 36.63 & 41.14 \\

L4DR~\cite{l4dr}
& 45.76 & \underline{26.63} & \textbf{26.18}
& \underline{21.59} & \underline{28.03} & \textbf{22.55}
& 43.02 & \underline{39.26} & \underline{43.70} \\

V2X-R~\cite{v2xr}
& 41.30 & 15.38 & 19.29
& 15.02 & 18.60 & 19.67
& \underline{45.06} & 25.79 & 36.80 \\

DCDA (Ours)
& \textbf{47.89} & \textbf{27.24} & \underline{25.23}
& \textbf{28.56} & \textbf{32.53} & \underline{20.73}
& \textbf{49.05} & \textbf{46.86} & \textbf{47.66} \\

\bottomrule
\end{tabular}
\end{table}

\subsection{Experimental Setup}
\label{sec:exp_setup}

\subsubsection{Dataset.}
\label{sec:dataset}

To evaluate the generalization capabilities of models beyond their training distributions, we establish a structured open-weather benchmark based on the K-Radar dataset. K-Radar provides synchronized LiDAR point clouds and 4D radar data across seven real-world conditions: Normal, Overcast, Fog, Rain, Sleet, LightSnow, and HeavySnow.
Building on these data, we define three evaluation protocols along the dimensions of weather type and severity. Detailed statistics and split implementation details are provided in Sec.~A of the supplementary material.
\begin{itemize}
\item \textbf{Type-open (real-world weather type holdout).}
We train on \emph{Normal} together with a subset of adverse weather types (\emph{seen} types), and evaluate on the full set of real-world conditions.
Adverse types not included in training (\emph{unseen} types) are encountered only at test time, directly measuring generalization to novel weather categories.
Concretely, we use the two most frequent adverse types in K-Radar---\textit{Rain} and \textit{Sleet}---as \emph{seen}, and hold out \textit{Overcast}, \textit{Fog}, \textit{LightSnow}, and \textit{HeavySnow}, which span mild to severe degradations not represented by the seen set.

\item \textbf{Severity-open (synthetic severity-level holdout).}
Severity variation is known to induce substantially different LiDAR degradation patterns even within the same weather type (e.g., fog/snow scattering)~\cite{Hahner2021FogSim,Hahner2022SnowSim}. Since K-Radar does not provide controlled severity annotations, we construct a synthetic severity setting from \emph{Normal} frames following Robo3D~\cite{robo3d}.
We consider three synthetic weather types (\emph{Fog, Snow, Wet Ground}) with three severity levels (\emph{Light/Medium/Heavy}).
Training observes only the \emph{Light} level for each synthetic type, while \emph{Medium} and \emph{Heavy} are held out for testing, evaluating robustness to unseen severity.
Although synthesis starts from Normal frames, DCDA does not exploit paired clean--corrupted supervision; weather labels are used only to define train/test splits.

\item \textbf{Type+severity-open.}
We further evaluate the most challenging intersection setting, where training observes only a single type--severity combination, and all remaining type--severity combinations are evaluated at test time.
\end{itemize}

\par\medskip\noindent\textbf{Metrics.}
Following the KITTI 3D detection protocol used in prior K-Radar works~\cite{Paek2022KRadar,l4dr},
we report BEV AP (\%) and 3D AP (\%) for the \emph{sedan} category at IoU $=0.5$.
Unless otherwise noted, all methods are evaluated using the same preprocessing and evaluation configuration.

\par\medskip\noindent\textbf{Implementation details.}
The detection-guided critic $\mathcal{H}$ is pretrained on the \textit{Normal}-only subset with a learning rate of $8\times10^{-3}$ and kept frozen thereafter.
The weather discriminator $\mathcal{D}$ is trained to distinguish \textit{Normal} from non-\textit{Normal} training samples with a learning rate of $5\times10^{-4}$, and is also frozen when training DCDA $\mathcal{A}_\theta$.
We optimize $\mathcal{A}_\theta$ with AdamW and a cosine-annealing schedule for $40$ epochs, using a batch size of $8$ and a learning rate of $1\times10^{-3}$, following the two-stage procedure in Sec.~\ref{sec:training_inference}.
We set the number of diffusion steps to $3$ for both training and inference, and use $(\lambda_1,\lambda_2)=(0.5,\,0.05)$ in Eq.~\eqref{eq:totalloss}.

\subsection{Main Results}
\label{sec:main_results}

To assess open-weather robustness for LiDAR--4D radar fusion 3D detection, we compare DCDA with representative baselines, including RTNH~\cite{Paek2022KRadar}, InterFusion~\cite{wang2022interfusion}, V2X-R~\cite{v2xr}, and L4DR~\cite{l4dr}, under the structured \emph{open-weather} protocols in Sec.~\ref{sec:dataset}.
The detailed results for each protocol are reported and analyzed in the following sections.

\begin{table}[t]
\centering
\caption{Results on the \textbf{type+severity-open}: BEV and 3D AP (\%) over held-out type--severity pairs. Training sees \emph{Normal} and \emph{Fog-Light} only.}
\label{tab:synthetic_type_severity_open}
\setlength{\tabcolsep}{3.6pt}
\scriptsize

\begin{tabular}{l ccc ccc ccc}
\toprule
\multirow{2}{*}{Method}
& \multicolumn{3}{c}{Fog}
& \multicolumn{3}{c}{Snow}
& \multicolumn{3}{c}{Wet Ground} \\
\cmidrule(lr){2-4}
\cmidrule(lr){5-7}
\cmidrule(lr){8-10}
& L & M & H
& L & M & H
& L & M & H \\
\midrule

\multicolumn{10}{c}{\textbf{BEV AP (\%)}} \\
\midrule

RTNH~\cite{Paek2022KRadar}
& 39.20 & 27.60 & 22.11
& 20.72 & 20.37 & 20.49
& 33.13 & 19.78 & 17.99 \\

InterFusion~\cite{wang2022interfusion}
& 58.09 & \underline{40.52} & 30.21
& 20.73 & 16.29 & 17.85
& 70.03 & 64.62 & \underline{66.97} \\

L4DR~\cite{l4dr}
& \underline{64.41} & \textbf{41.03} & \underline{31.49}
& \underline{21.43} & 21.88 & \underline{22.36}
& \underline{72.16} & 64.55 & \textbf{69.12} \\

V2X-R~\cite{v2xr}
& 62.89 & 38.46 & \textbf{32.79}
& 20.99 & \underline{22.96} & \textbf{23.83}
& 71.37 & \underline{64.78} & 64.91 \\

DCDA (Ours)
& \textbf{68.68} & 39.15 & 28.65
& \textbf{25.54} & \textbf{28.15} & 19.17
& \textbf{74.09} & \textbf{65.78} & 66.24 \\

\midrule
\multicolumn{10}{c}{\textbf{3D AP (\%)}} \\
\midrule

RTNH~\cite{Paek2022KRadar}
& 10.50 & 12.39 & 10.21
& \underline{14.53} & 8.90 & 11.77
& 13.12 & 13.64 & 12.20 \\

InterFusion~\cite{wang2022interfusion}
& 39.64 & \underline{24.36} & 18.75
& 11.16 & 14.49 & \underline{15.89}
& 39.89 & \underline{40.15} & 37.44 \\

L4DR~\cite{l4dr}
& 43.75 & 23.41 & \textbf{23.53}
& 11.72 & 13.74 & 10.88
& \underline{49.59} & 37.25 & \textbf{42.44} \\

V2X-R~\cite{v2xr}
& \underline{44.21} & 23.50 & \underline{22.11}
& 11.87 & \underline{18.36} & \textbf{15.91}
& 44.88 & 38.85 & 40.46 \\

DCDA (Ours)
& \textbf{45.38} & \textbf{27.25} & 19.83
& \textbf{16.27} & \textbf{19.48} & 12.46
& \textbf{51.92} & \textbf{41.15} & \underline{41.62} \\

\bottomrule
\end{tabular}
\end{table}

\par\medskip\noindent\textbf{Type-open (real weather types).}
To evaluate generalization to entirely unseen real-world weather types, we first conduct experiments under the type-open protocol.
Table~\ref{tab:kradar_type_open} shows that while strong baselines (e.g., L4DR) remain competitive on the seen split, their performance degrades noticeably on unseen adverse types (e.g., \textit{Fog} and \textit{HeavySnow} in our split).
In contrast, DCDA improves robustness on unseen types and reduces the seen--unseen gap.
Averaged over all unseen weather types, DCDA surpasses the best baseline (L4DR) by \textbf{3.83} BEV AP and \textbf{7.73} 3D AP, while maintaining comparable accuracy on the seen split.
Furthermore, to examine robustness under different evaluation operating points, we sweep IoU thresholds over $\{0.3,0.5,0.7\}$ and score thresholds over $\{0.1,0.2,0.3\}$. Across all nine combinations, DCDA has the highest \emph{mean} unseen-weather BEV AP, leading the best baseline by \textbf{1.6} to \textbf{7.3} points. The largest gain is observed under the strict IoU\,$=\,0.7$ setting, and the full grid is provided in Sec.~C.1 of the supplementary material.

\par\medskip\noindent\textbf{Severity-open (synthetic severity).}
To evaluate robustness to unseen severity levels within known synthetic weather types, we train on \textit{Normal} and \textit{Light}, and evaluate on \textit{Light} and held-out \textit{Medium/Heavy} levels.
As reported in Table~\ref{tab:synthetic_severity_open}, DCDA achieves the strongest mean performance over the unseen combinations, improving unseen BEV/3D AP by \textbf{1.28}/\textbf{2.32} percentage points over L4DR.
Across \textit{Fog}, \textit{Snow}, and \textit{Wet Ground}, DCDA stays competitive as degradation increases.
Under the most severe cases (e.g., \textit{Heavy Snow}), performance remains challenging for all methods due to extreme LiDAR corruption, but DCDA still provides the best overall robustness trend.

\par\medskip\noindent\textbf{Type+severity-open (synthetic type \& severity).}
To evaluate the most challenging joint holdout where both weather type and severity may be unseen at test time,
we train only on \textit{Normal} and a single adverse configuration (\textit{Fog-Light}), and evaluate on all type--severity combinations.
As reported in Table~\ref{tab:synthetic_type_severity_open}, despite the extremely limited adverse exposure, DCDA attains the highest mean unseen BEV AP and the highest mean unseen 3D AP,
exceeding V2X-R by \textbf{1.76} percentage points and L4DR by \textbf{2.18} percentage points on mean unseen 3D AP.
These results suggest DCDA can transfer beyond the single observed adverse setting.

\begin{table}[t]
\centering
\caption{Synthetic-trained models evaluated on real K-Radar weather (BEV/3D AP \%). Non-\textit{Normal} denotes the macro mean over the six adverse conditions.}
\label{tab:real_weather}
\setlength{\tabcolsep}{5pt}
\scriptsize

\begin{tabular}{l cc cc}
\toprule
\multirow{2}{*}{Method}
& \multicolumn{2}{c}{Severity-open}
& \multicolumn{2}{c}{Type+severity-open} \\
\cmidrule(lr){2-3}\cmidrule(lr){4-5}
& Normal & non-Normal & Normal & non-Normal \\
\midrule
InterFusion & $67.9/41.7$ & $53.5/28.2$ & $66.9/41.3$ & $45.9/18.1$ \\
V2X-R       & $66.5/35.2$ & $55.5/20.5$ & $65.6/42.2$ & $49.8/23.3$ \\
L4DR        & $67.9/\mathbf{49.1}$ & $56.1/\mathbf{30.0}$ & $\mathbf{67.3}/\mathbf{43.4}$ & $50.6/25.5$ \\
DCDA (Ours) & $\mathbf{73.3}/48.2$ & $\mathbf{58.8}/27.3$ & $67.1/\mathbf{43.4}$ & $\mathbf{54.2}/\mathbf{26.8}$ \\
\bottomrule
\end{tabular}%

\end{table}

\begin{table}[t]
\centering
\caption{Real \textit{LightSnow}$\rightarrow$\textit{HeavySnow} severity transfer (BEV/3D AP \%): models trained on real \textit{Normal}\,$+$\,\textit{LightSnow} and tested on real \textit{Normal}\,$+$\,\textit{HeavySnow}.}
\label{tab:ls2hs}
\setlength{\tabcolsep}{5pt}
\scriptsize

\begin{tabular}{l cc}
\toprule
Method & Normal & HeavySnow \\
\midrule
L4DR & $\mathbf{71.61}/\mathbf{51.20}$ & $30.57/16.57$ \\
DCDA (Ours) & $68.39/50.01$ & $\mathbf{37.92}/\mathbf{25.61}$ \\
\bottomrule
\end{tabular}%

\end{table}

\par\medskip\noindent\textbf{Generalization to real weather.}
Since the severity-open and type+severity-open evaluations rely on \emph{synthetic} weather, we further test DCDA on \emph{real} adverse weather through the following two experiments on K-Radar:
(1) \emph{synthetic-to-real direct testing}, where models trained under the synthetic protocols are evaluated on real K-Radar weather conditions without further training;
and (2) \emph{real LightSnow-to-HeavySnow generalization}, where models are trained on real light snow and tested on real heavy snow.
\begin{itemize}
\item\textbf{Synthetic-to-real direct testing.}
To examine whether models trained with synthetic adverse weather can generalize to real weather conditions, we take the models already trained under the severity-open and type+severity-open protocols and evaluate them, as is, on the seven real K-Radar conditions.
Table~\ref{tab:real_weather} reports the averaged results, and the per-condition results are provided in Sec.~C.2 of the supplementary material.
Averaged over Rain, Sleet, Fog, Overcast, LightSnow, and HeavySnow, DCDA achieves the best BEV AP under both protocols, exceeding L4DR by \textbf{2.7} AP in the severity-open setting and by \textbf{3.6} AP in the type+severity-open setting, while maintaining competitive \textit{Normal} accuracy.

\item\textbf{Real LightSnow-to-HeavySnow generalization.}
To further assess generalization across real severity levels, we train the models on real \textit{Normal}\,$+$\,\textit{LightSnow} and test them on real \textit{Normal}\,$+$\,\textit{HeavySnow}.
As reported in Table~\ref{tab:ls2hs}, DCDA improves this challenging transfer case over L4DR by \textbf{7.35} BEV AP and \textbf{9.04} 3D AP on \textit{HeavySnow}.
\end{itemize}

\begin{table}[t]
\centering
\caption{Ablation study on K-Radar \textbf{type-open}: mean BEV and 3D AP (\%) on seen and unseen weather types.}
\scriptsize
\setlength{\tabcolsep}{6pt}
\renewcommand{\arraystretch}{1.1}

\begin{tabular}{ccc|cccc}
\toprule
$\mathcal{L}_{\text{diff}}$ 
&$\mathcal{L}_{\text{det}}$ 
&$\mathcal{L}_{\text{adv}}$ 
& Seen BEV 
& Seen 3D 
& Unseen BEV 
& Unseen 3D \\
\midrule

\checkmark &  &  & 61.76 & 43.12 & 66.23 & 39.11 \\

\checkmark & \checkmark &  & 66.68 & 44.95 & 70.47 & 41.50 \\

\checkmark &  & \checkmark & 63.97 & 44.06 & 68.22 & 40.99 \\

\checkmark & \checkmark & \checkmark 
& 66.77 & 45.47 & 72.65 & 42.25 \\

\bottomrule
\end{tabular}
\label{tab:ablation_components}
\end{table}

\subsection{Ablation Study}
\label{sec:ablation}
We perform ablation studies to analyze the contribution of each component in DCDA.
Table~\ref{tab:ablation_components} reports the results under the \textit{type-open} protocol.
Using structural diffusion alone ($\mathcal{L}_{\text{diff}}$) yields limited cross-weather robustness. Integrating the detection-guided critic ($\mathcal{L}_{\text{det}}$) significantly boosts performance, confirming the necessity of task-level semantic constraints. 
Adding the weather adversarial critic ($\mathcal{L}_{\text{adv}}$) encourages distributional alignment, further improving unseen-weather generalization.
Adding both critics to the diffusion objective achieves the highest AP, showing that semantic preservation and distributional alignment are complementary for recovering a clean feature manifold.

\begin{figure}[t]
    \centering
    \includegraphics[width=0.95\linewidth,height=0.14\textheight]{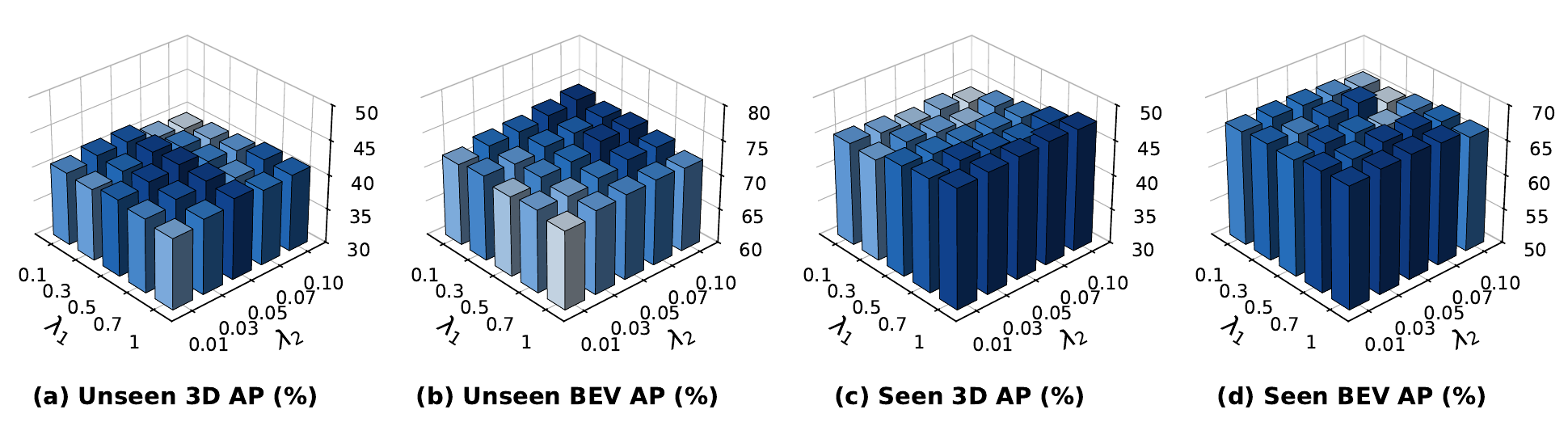}
    \caption{Parameter sensitivity analysis on K-Radar \textbf{type-open}: mean BEV/3D AP (\%) on seen and unseen weather types.}
    \label{fig:lambda_sensitivity}
\end{figure}

\subsection{Other Experiments}

\par\medskip\noindent\textbf{Parameter Sensitivity.}
We evaluate DCDA's sensitivity to the critic weights $\lambda_1$ and $\lambda_2$ in Eq.~\eqref{eq:totalloss} under the type-open protocol.
We sweep $\lambda_1 \in \{0.1,\allowbreak 0.3,\allowbreak 0.5,\allowbreak 0.7,\allowbreak 1.0\}$ and $\lambda_2 \in \{0.01, 0.03, 0.05, 0.07, 0.1\}$ around the default $(0.5,\,0.05)$.
Fig.~\ref{fig:lambda_sensitivity} shows that DCDA remains stable across a broad hyperparameter neighborhood, suggesting that its dual-critic guidance is robust to moderate weight variations.

\begin{table}[t]
\centering
\caption{Robustness--efficiency trade-off on K-Radar \textbf{type-open} (BEV AP \%). Latency (ms), throughput (FPS), and peak memory (MiB) measured on a single GPU.}
\label{tab:efficiency}
\setlength{\tabcolsep}{4pt}
\scriptsize

\begin{tabular}{l cccc ccc}
\toprule
Method & Normal & Seen & Unseen & All & ms & FPS & MiB \\
\midrule
w/o DCDA              & 68.41 & 64.28 & 65.96 & 65.24 & \textbf{29.51} & \textbf{33.89} & \textbf{470.7} \\
+\,DCDA               & \textbf{68.43} & 66.87 & \textbf{72.65} & \textbf{70.13} & 70.86 & 14.11 & 573.7 \\
+\,DCDA\,+\,routing   & 68.39 & \textbf{66.77} & \underline{70.90} & \underline{69.13} & \underline{59.59} & \underline{16.78} & 710.1 \\
\bottomrule
\end{tabular}%

\end{table}

\par\medskip\noindent\textbf{Efficiency and Routing.}
We further analyze the robustness--efficiency trade-off of DCDA under the type-open protocol, with results reported in Table~\ref{tab:efficiency}. Applying DCDA to every frame yields the largest robustness gain of \textbf{6.69} unseen BEV AP, increasing latency from $29.51$ to $70.86$\,ms.
The routed variant skips DCDA for frames classified as \textit{Normal} with high confidence by the discriminator $\mathcal{D}$.
With $99.8\%$ routing accuracy, it retains most of the robustness gain, improving unseen BEV AP by \textbf{4.94} points while reducing latency to $59.59$\,ms.

\par\medskip\noindent\textbf{Capacity Control.}
To isolate the effect of diffusion refinement from model capacity, we conduct a capacity-controlled comparison by replacing the diffusion aligner with a same-budget GAN refiner.
Table~\ref{tab:refiner_comparison} shows that DCDA exceeds this refiner by \textbf{6.92} BEV AP, indicating that the gain comes from coupling diffusion refinement with dual-critic guidance rather than added capacity.

\begin{table}[t]
\centering
\caption{Same-detector and same-budget generative-refiner comparison on K-Radar \textbf{type-open}. BEV/3D AP~(\%).}
\label{tab:refiner_comparison}
\setlength{\tabcolsep}{5pt}
\scriptsize

\begin{tabular}{l cc}
\toprule
Refiner & Seen BEV/3D & Unseen BEV/3D \\
\midrule
GAN refiner & $61.40/43.01$ & $65.73/40.52$ \\
DCDA (Ours) & $\mathbf{66.77}/\mathbf{45.47}$ & $\mathbf{72.65}/\mathbf{42.25}$ \\
\bottomrule
\end{tabular}%
\end{table}

\begin{figure}[!b]
    \centering
    \includegraphics[width=\linewidth]{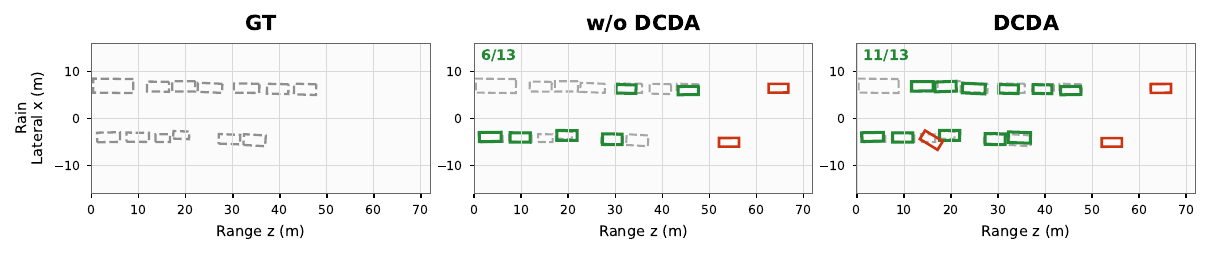}
    \caption{Qualitative BEV detections on a real K-Radar \textit{Rain} scene (\textbf{type-open}). Columns: ground truth (dashed gray), w/o DCDA, and DCDA (Ours). Green: matched predictions; red: false positives. Numbers: matched-GT out of total GT objects.}
    \label{fig:qualitative_detection}
    \medskip
    \centering
    \includegraphics[width=\linewidth,
        height=0.20\textheight]{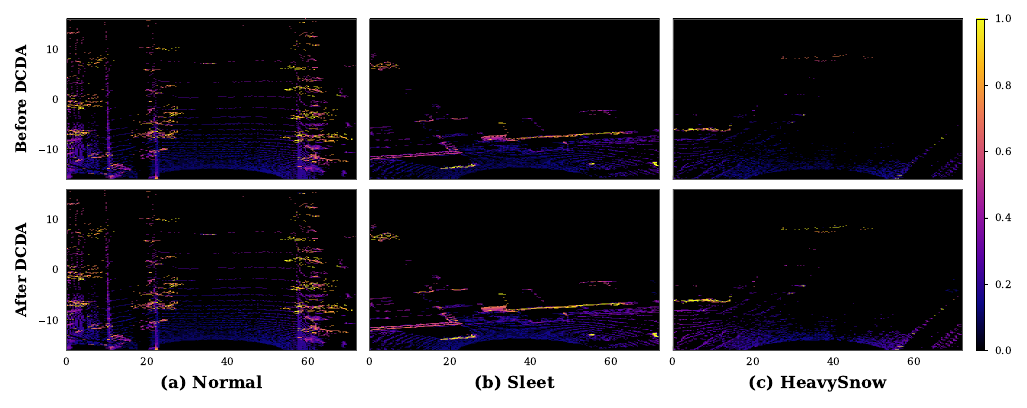}
    \caption{Visualization on K-Radar \textbf{type-open}: intermediate BEV feature maps before (top) and after (bottom) applying DCDA under (a) \textit{Normal}, (b) \textit{Sleet}, (c) \textit{HeavySnow} weather conditions.}
    
    \label{fig:bev_feature}
\end{figure}

\par\medskip\noindent\textbf{Visualization Results.}
\label{sec:visualization}
To qualitatively understand how DCDA helps under adverse weather, we visualize its effect at both the detection and feature levels. 
At the detection level, Fig.~\ref{fig:qualitative_detection} shows BEV detections on a real \textit{Rain} scene, where DCDA correctly matches 11 of the 13 ground-truth objects, versus only 6 for the baseline. 
At the feature level, Fig.~\ref{fig:bev_feature} shows that DCDA reshapes the intermediate LiDAR BEV feature maps by strengthening weak object-related responses and partially recovering suppressed evidence, while preserving the LiDAR-centric layout.

\subsection{Limitations}
\label{sec:limitations}
While DCDA delivers consistent open-weather gains, several limitations remain and point to future work. First, iterative refinement adds inference overhead, leaving a speed--robustness trade-off that routing alleviates but does not eliminate.
Second, DCDA's added margin is smaller at the most extreme severities (e.g., \textit{Snow-Heavy}, \textit{Fog-Heavy}), where LiDAR geometry is largely degraded and little structure remains for the anchored refinement to recover---a regime that is intrinsically challenging for most fusion methods.
Finally, our controlled severity evaluation is synthetic and DCDA currently assumes \textit{Normal} as the clean reference; extending to richer real-severity annotations and heterogeneous clean references is future work.

\section{Conclusion}
\label{sec:conclusion}
We investigate open-weather generalization for LiDAR--4D radar fusion 3D detection, where test-time weather may differ from training conditions in both type and severity. We propose DCDA, a weather-agnostic diffusion alignment framework that restores degraded LiDAR features through 4D radar-guided refinement with complementary semantic and distributional supervision. Without paired clean--degraded data or weather type/severity labels as conditional inputs, DCDA consistently improves robustness under unseen weather while remaining competitive on seen conditions. We hope this work provides a practical step toward robust and generalizable multi-modal 3D perception in real-world weather.



\section*{Acknowledgements}
This research was supported in part by the National Natural Science Foundation of China under Grant 62576167, the Young Scientists Fund of the National Natural Science Foundation of China under Grant 62402220, and the Natural Science Foundation of Jiangsu Province under Grant BK20241402.

\newpage
%
%
\bibliographystyle{splncs04}
\bibliography{main}
\end{document}